\documentclass{article}

\usepackage{arxiv}

\usepackage[utf8]{inputenc} 
\usepackage[T1]{fontenc}    
\usepackage{hyperref}       
\usepackage{url}            
\usepackage{booktabs}       
\usepackage{amsfonts}       
\usepackage{amsmath}
\usepackage{nicefrac}       
\usepackage{microtype}      
\usepackage{subcaption}
\usepackage{cleveref}       
\usepackage{lipsum}         
\usepackage{graphicx}
\usepackage{natbib}
\usepackage{doi}

\usepackage{textcomp}
\usepackage{array}
\usepackage{floatrow}

\title{Embedding And Clustering Your Data Can Improve Contrastive Pretraining}

\newif\ifuniqueAffiliation
\uniqueAffiliationtrue

\ifuniqueAffiliation 
\author{ Luke Merrick\thanks{\url{https://lukemerrick.com}} \\
	Snowflake Inc. \\
	\texttt{luke.merrick@snowflake.com} \\
}
\else
\usepackage{authblk}

\newbox{\orcid}\sbox{\orcid}{\includegraphics[scale=0.06]{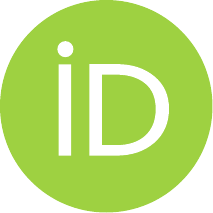}} 
\author[1]{%
	\href{https://orcid.org/0000-0000-0000-0000}{\usebox{\orcid}\hspace{1mm}David S.~Hippocampus\thanks{\texttt{hippo@cs.cranberry-lemon.edu}}}%
}
\author[1,2]{%
	\href{https://orcid.org/0000-0000-0000-0000}{\usebox{\orcid}\hspace{1mm}Elias D.~Striatum\thanks{\texttt{stariate@ee.mount-sheikh.edu}}}%
}
\affil[1]{Department of Computer Science, Cranberry-Lemon University, Pittsburgh, PA 15213}
\affil[2]{Department of Electrical Engineering, Mount-Sheikh University, Santa Narimana, Levand}
\fi

\renewcommand{\shorttitle}{Embedding And Clustering Your Data Can Improve Contrastive Pretraining}

\hypersetup{
pdftitle={Embedding And Clustering Your Data Can Improve Contrastive Pretraining},
pdfauthor={Luke Merrick},
pdfkeywords={Contrastive learning, Information retrieval, Text embedding},
}

\begin{document}
\maketitle

\begin{abstract}
	Recent studies of large-scale contrastive pretraining in the text embedding domain show that using single-source minibatches, rather than mixed-source minibatches, can substantially improve overall model accuracy. In this work, we explore extending training data stratification beyond source granularity by leveraging a pretrained text embedding model and the classic $k$-means clustering algorithm to further split training data apart by the semantic clusters within each source. Experimentally, we observe a notable increase in NDCG@10 when pretraining a BERT-based text embedding model on query-passage pairs from the MSMARCO passage retrieval dataset. Additionally, we conceptually connect our clustering approach to both the Topic Aware Sampling (TAS) aspect of the TAS-B methodology and the nearest-neighbor-based hard-negative mining aspect of the ANCE methodology and discuss how this unified view motivates future lines of research on the organization of contrastive pretraining data.
\end{abstract}


\section{Introduction}\label{sec:intro}

\begin{figure}
\floatbox[{\capbeside\thisfloatsetup{capbesideposition={right,center},capbesidewidth=6cm}}]{figure}[\FBwidth]
{
	\caption{Figure 7 from \citep{merrick2024arctic}, which depicts the NDCG@10 score on the SciDocs dataset during training. This experiment showed that stratified training data (dark blue line) can lead to better quality than simply a large batch size (purple line), while a combination of both approaches (light blue line) does even better than each individual. It also shows how the stratified approach improves the long-term trajectory more, suggesting an element of curriculum learning may be at play. }
	\label{fig:copied_figure}
}
{
    \includegraphics[width=8cm]{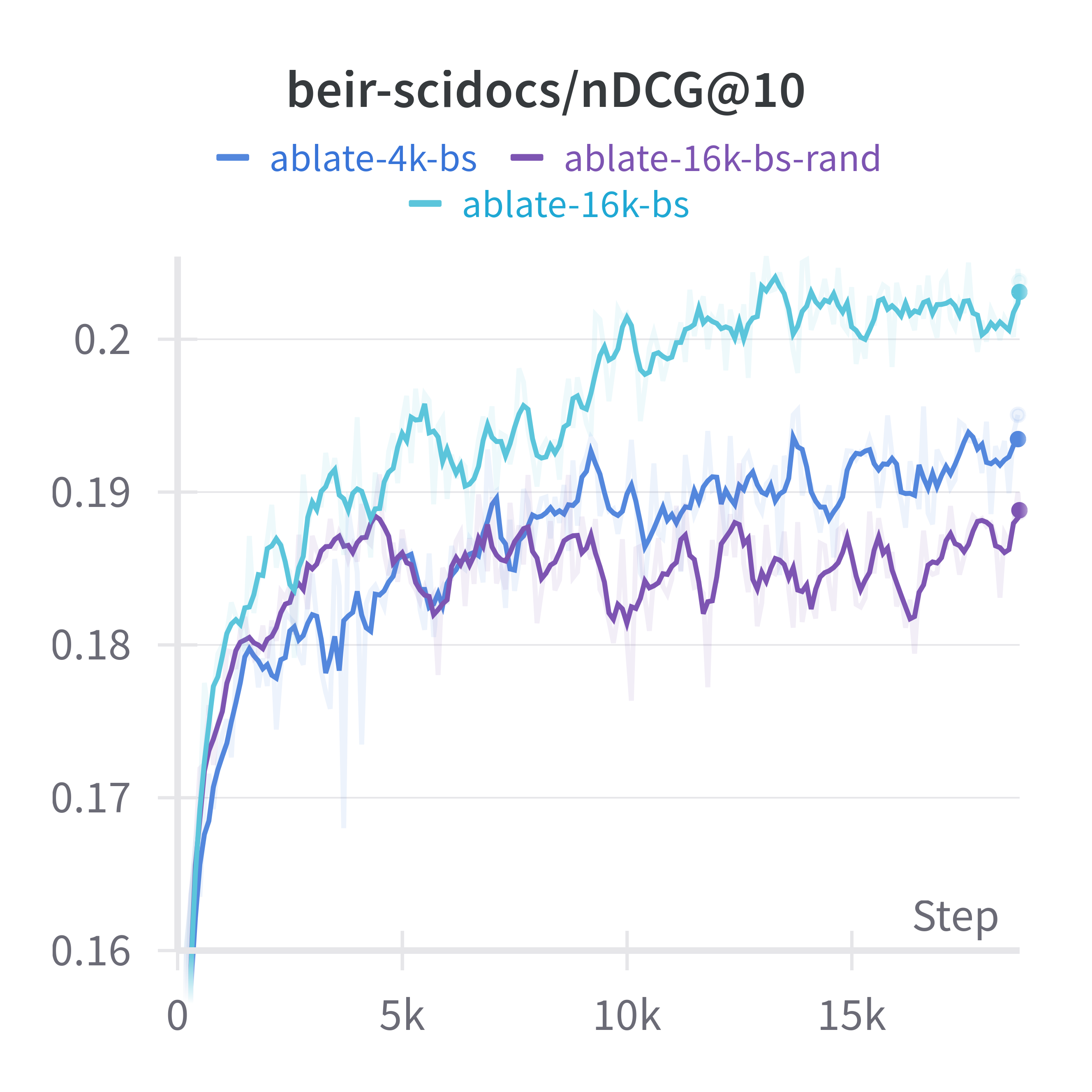}
}
\end{figure}

The Snowflake Arctic Embed report \citep{merrick2024arctic} showed that pretraining with stratified data sampling that fills each minibatch with samples from just a single source can lead to better model performance than the simpler approach of randomly shuffling all data sources together into a single mixed dataset (see \Cref{fig:copied_figure}, which is reproduced from that work). The Nomic-Embed report \citep{nussbaum2024nomic} also cites this same trick as part of their successful recipe for training, claiming it prevents the model from ``learning source-specific shortcuts.''

Observing the empirical retrieval quality improvement in the Arctic Embed report's ablation study, two complimentary research objectives naturally arise:

\begin{enumerate}
    \item \textbf{Applied objective.} Can we extend the source stratification method to further improve large-scale contrastive pretraining learning dynamics? \label{ro:applied}
    \item \textbf{Theoretical objective.} Can we develop a deeper theoretical understanding of how and why source stratification drives these learning improvements?\label{ro:theory}
\end{enumerate}

In pursuit of these goals, we concoct and study a simple extension of source stratification that creates more granular ``semantic sub-sources'' by clustering examples by their text embeddings. Empirically, our experiments using the MSMARCO dataset \citep{msmarco} show a clear improvement from this extended stratification technique, while theoretically, it motivates conceptual connection to prior works on Task Aware Sampling (TAS) \citep{tasb2021} and nearest-neighbors-based hard negative mining \citep{xiong2020approximatenearestneighbornegative}. Though we fall short of a conclusive theoretical explanation, we offer an initial synthesis of relevant theoretical principles to provide a cohesive picture of improved contrastive pretraining at an intuitive level. Additionally, we believe our synthesized viewpoint offers insight into promising future lines of related research, which we discuss at the close of this paper.

\section{Methodology}\label{sec:methodology}

Our methodology for extending source stratification to a more granular level is a rather straightforward application of clustering and is closely related to the Topic Aware Sampling (TAS) approach of \citep{tasb2021}.

Our clustering recipe requires two input objects:
\begin{itemize}
    \item A large-scale contrastive pretraining dataset of $N$ query-item pairs, i.e. $D = \{q_i, p_i\}_{i=1}^N$.
    \item A pretrained text embedding model $f$ mapping queries and items to the same vector space.
\end{itemize}

Using these objects, we apply the following steps:
\begin{itemize}
    \item Use model $f$ to create a set of $N$ pair embeddings, either by embedding all the queries or all the items\footnote{Developing a way to combine query and item into a single embedding vector presents an interesting line for future work as well.}, i.e. $Z = \{f(q_i)\}_{i=1}^N$ or $Z = \{f(p_i)\}_{i=1}^N$.
    \item Perform $k$-means clustering to assign each embedding in $Z$ to one of $k$ clusters.
    \item Construct datasets $D_1, D_2, \ldots, D_k$ by splitting the original query-item pairs of $D$ into sub-datasets corresponding to the cluster assignments of the items' embeddings.
    \item Stratify across $D_1, D_2, \ldots, D_k$, rather than $D$, when constructing source-stratified minibatches for largescale contrastive pretraining.
\end{itemize}

\section{Experiments}

To assess the practical efficacy of our clustering method, we adapt the MSMARCO passage dataset \citep{msmarco} to the setting of large-scale contrastive pretraining by discarding all labeled negative examples, leaving only a set of query-passage pairs. Although MSMARCO contains over 8 million passages, there are only about 0.5 million labeled query-passage pairs in the training set, so our dataset ends up being not particularly ``large scale''.

After converting the original MSMARCO passage dataset to a pretraining-compatible dataset, we perform embedding and clustering on queries and vectors separately to create two stratified training datasets. We then perform contrastive pretraining on the original dataset and the two stratified variants and evaluate the quality of the resulting models.

\subsection{Clustering Details}

We begin the clustering process by embedding both queries and passages using the Arctic Embed M model \citep{merrick2024arctic}. We then leverage the spherical $k$-means clustering algorithm implemented in the FAISS library \citep{douze2024faiss} with $k=10$ clusters\footnote{We arbitrarily select $k=10$ since it is a round number in the ballpark of the number of different data sources used by the Snowflake Arctic Embed and Nomic-Embed projects.} to construct our two clustered datasets. Cluster statistics are given in \Cref{tab:clusters}.

\begin{table}
    \small
	\caption{Cluster statistics. To give a rough approximation of cluster density, we sample 3,000 points from both the overall dataset and each cluster and compute the average pairwise cosine similarity within each cluster.}
    \begin{subtable}[t]{0.45\textwidth}
	\caption{Passage embedding cluster statistics.}
	\centering
	\begin{tabular}{lll}
		\toprule
		Cluster & Size & Similarity \\
		\midrule
		000 & 61,655  & 0.340   \\
		001 & 68,323  & 0.358   \\
		002 & 30,173  & 0.350   \\
		003 & 44,481  & 0.422   \\
		004 & 40,552  & 0.387   \\
		005 & 34,756  & 0.290   \\
		006 & 36,700  & 0.263   \\
		007 & 90,862  & 0.493   \\
		008 & 74,908  & 0.438   \\
		009 & 34,062  & 0.360   \\
        \midrule
        Overall Dataset &  & 0.305 \\
		\bottomrule
        \label{tab:clusters_doc}
	\end{tabular}
    \end{subtable}
    \hspace{\fill}
    \begin{subtable}[t]{0.45\textwidth}
	\caption{Query embedding cluster statistics.}
	\centering
	\begin{tabular}{lll}
		\toprule
		Cluster & Size & Similarity \\
		\midrule
		001 & 51,475 &  0.705 \\
        002 & 33,574 &  0.633 \\
        003 & 67,224 &  0.769 \\
        004 & 42,233 &  0.711 \\
        005 & 35,114 &  0.699 \\
        006 & 79,377 &  0.729 \\
        007 & 46,528 &  0.746 \\
        008 & 61,172 &  0.715 \\
        009 & 26,574 &  0.682 \\
        010 & 59,668 &  0.703 \\
        \midrule
        Overall Dataset &  & 0.670 \\
		\bottomrule
	\end{tabular}
    \end{subtable}
	\label{tab:clusters}
\end{table}

\subsection{Training}

\begin{figure}
	\centering
    \includegraphics[width=0.7\linewidth]{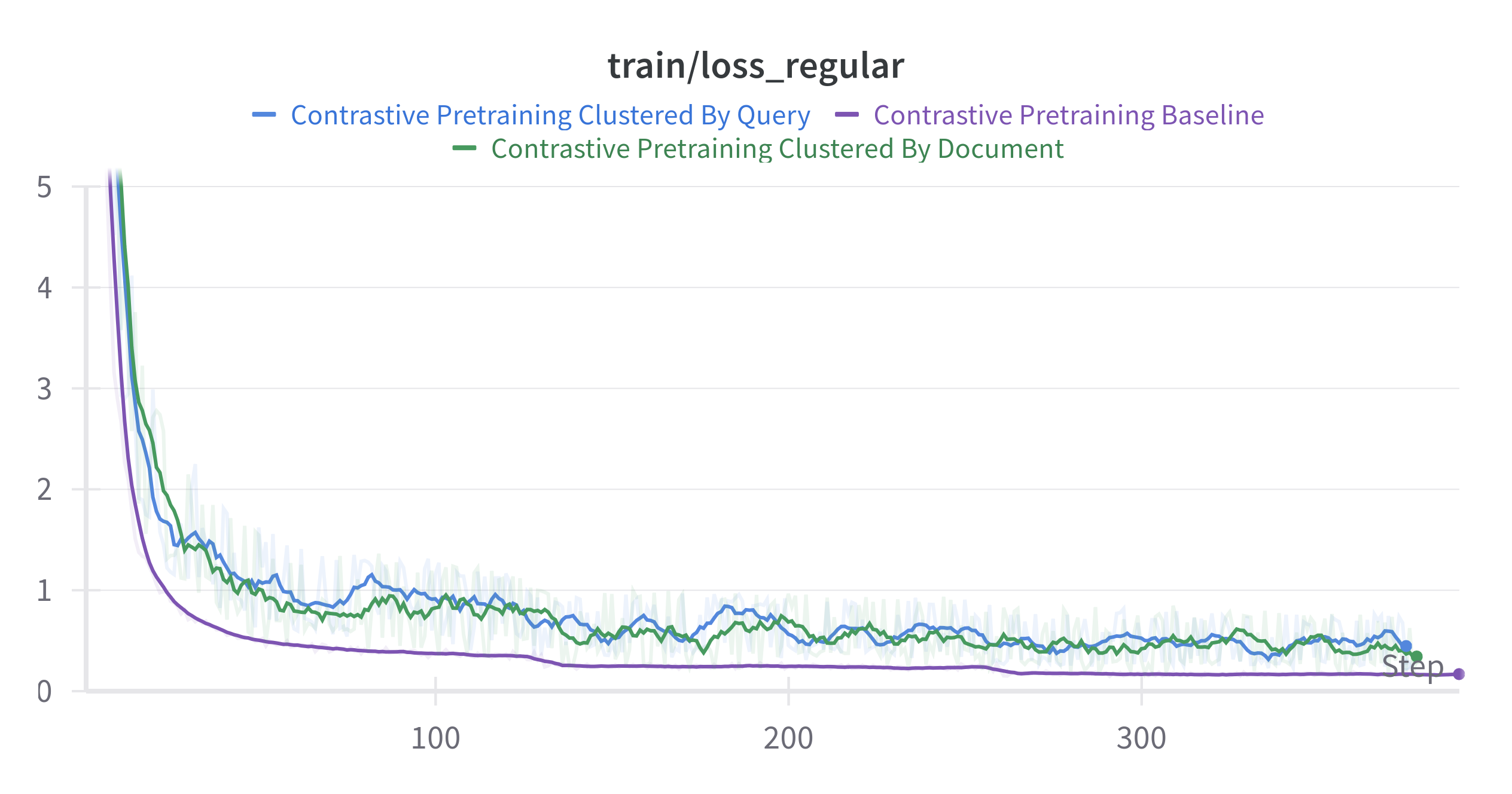}
	\caption{Experimental training loss curves (rolling average of 10 steps over faded original values). Clustering by pseudo-sub-sources leads to substantially higher average training loss, as well as higher variance step-to-step. }
	\label{fig:train_loss}
\end{figure}

We train a base-sized BERT model \citep{devlin2019bert} using \texttt{[CLS]}-token-based embeddings. From \Cref{fig:train_loss}, we see that compared to the un-clustered baseline, the clustered treatments lead to substantially higher training loss (i.e. increased negative hardness). Training parameters are given in \Cref{sec:training_parameters}.

\subsection{Results}

We evaluate our two pre-trained models on the full MTEB Retrieval benchmark \citep{muennighoff2023mtebmassivetextembedding} (which includes the dev split of MSMARCO as one of its constituent datasets) and tabulate the results in \Cref{tab:results}. We observe that our $k=10$ clustering leads to roughly 2\% improvement in NDCG@10 score on the MSMARCO dev split both when clustering by query and by passage embeddings, validating our hypothesis that further stratification would lead to improved performance.

The overall MTEB Retrieval evaluation paints a more nuanced picture than the MSMARCO dev split score, however. While passage-driven clustering is associated with improvements on 11 out of the 15 datasets, it demonstrates significant losses on ClimateFEVER, SCIDOCS, and SciFact. Interestingly, we see that the score degradation on these three datasets, along with the strong improvement on FiQA, are mirrored when training with clustered stratification driven by query embeddings, though performance elsewhere is more mixed. After examining some cluster examples manually (see \Cref{sec:cluster_examples}, in particular, passage cluster 3), we conjecture that clustering led to some minibatches much closer resembling the distribution of the FiQA dataset and thus helping the model learn to perform well in that domain.

Though these experiments only show a 0.7\% increase in MTEB Retrieval score using the better of the two variations considered, we contend that the utility of clustering may be more significant on larger datasets that can support a larger number of bigger-than-batch-size clusters. Although we leave a formal ablation study at the 100M+ query-item pair scale to future work, anecdotally we find that clustering by passage embedding at the 100M+ passage scale into hundreds of clusters appears to be associated with score improvements as high as a couple percent in long pretraining runs (e.g. ballpark 48\% NDCG@10 on MTEB Retrieval instead of the ballpark 47\% published in the ablation study of \citep{merrick2024arctic}).


\begin{table}
	\caption{Retrieval results on the MTEB Retrieval benchmark. Scores in NDCG@10. In-distribution dataset (MSMARCO) bolded.}
    \small
    \begin{subtable}[t]{0.48\textwidth}
	\centering
	\caption{Raw NDCG@10 scores.}
    \begin{tabular}{lrrr}
    \toprule
    Dataset & Baseline & Doc Cluster & Query Cluster \\
    \midrule
    ArguAna & 44.98\% & 45.72\% & 46.06\% \\
    CQADup. & 30.53\% & 31.24\% & 31.60\% \\
    ClimateFEVER & 20.35\% & 19.72\% & 19.37\% \\
    DBPedia & 31.49\% & 31.61\% & 32.41\% \\
    FEVER & 66.02\% & 67.38\% & 64.72\% \\
    FiQA2018 & 24.05\% & 25.15\% & 25.14\% \\
    HotpotQA & 50.79\% & 50.94\% & 50.45\% \\
    \textbf{MSMARCO} & \textbf{32.86\%} & \textbf{33.58\%} & \textbf{33.48\%} \\
    NFCorpus & 28.30\% & 28.53\% & 28.62\% \\
    NQ & 37.04\% & 37.26\% & 38.30\% \\
    QuoraRetrieval & 84.66\% & 84.78\% & 84.93\% \\
    SCIDOCS & 14.05\% & 13.76\% & 13.79\% \\
    SciFact & 57.70\% & 55.75\% & 56.74\% \\
    TRECCOVID & 48.40\% & 49.93\% & 47.19\% \\
    Touche2020 & 17.24\% & 17.20\% & 16.45\% \\
    \midrule
    Average & 39.23\% & 39.50\% & 39.28\% \\
    \bottomrule
    \end{tabular}
    \end{subtable}
    \hspace{\fill}
    \begin{subtable}[t]{0.48\textwidth}
	\centering
	\caption{NDCG@10 scores normalized to shuffle.}
    \begin{tabular}{lrrr}
    \toprule
    Dataset & Baseline & Doc Cluster & Query Cluster \\
    \midrule
    ArguAna & 100.00\% & 101.65\% & 102.40\% \\
    CQADup. & 100.00\% & 102.32\% & 103.50\% \\
    ClimateFEVER & 100.00\% & 96.90\% & 95.18\% \\
    DBPedia & 100.00\% & 100.38\% & 102.92\% \\
    FEVER & 100.00\% & 102.06\% & 98.03\% \\
    FiQA2018 & 100.00\% & 104.57\% & 104.53\% \\
    HotpotQA & 100.00\% & 100.30\% & 99.33\% \\
    \textbf{MSMARCO} & \textbf{100.00\%} & \textbf{102.19\%} & \textbf{101.89\%} \\
    NFCorpus & 100.00\% & 100.81\% & 101.13\% \\
    NQ & 100.00\% & 100.59\% & 103.40\% \\
    QuoraRetrieval & 100.00\% & 100.14\% & 100.32\% \\
    SCIDOCS & 100.00\% & 97.94\% & 98.15\% \\
    SciFact & 100.00\% & 96.62\% & 98.34\% \\
    TRECCOVID & 100.00\% & 103.16\% & 97.50\% \\
    Touche2020 & 100.00\% & 99.77\% & 95.42\% \\
    \midrule
    Average & 100.00\% & 100.69\% & 100.13\% \\
    \bottomrule
    \end{tabular}
    \end{subtable}
    \label{tab:results}
\end{table}

\section{In Search Of Deeper Understanding}

In this section we turn our attention from our applied research objective to our theoretical research objective, connecting our empirical observation above to the concepts presented in several related works. After making these connections, we synthesize the ideas from these different sources into a unified perspective on how to more effectively organize contrastive pretraining data into minibatches.

\subsection{Concept One: The Clustering Hypothesis And Topicality Aware Sampling}\label{sec:cluster_hypothesis}

Topic Aware Sampling (TAS) -- an approach consisting of clustering training data by query embedding and then sampling each training minibatch from a single cluster each, was one of the key drivers of success for TAS-B \citep{tasb2021}, a holistic training recipe designed to produce a quality text embedding model without requiring large-scale computational resources. While at first, this technique may sound identical to the method proposed by this paper, the two differ somewhat. TAS calls for a dataset containing many labeled negative examples per query, i.e. the kind used in the finetuning stage of works like E5 \cite{wang2024e5} and Arctic Embed \cite{merrick2024arctic}, and it is thus not directly applicable to large-scale query-item pair datasets like those used in the pretraining step of these more recent text embedding models.

However, it is certainly fair to view the clustering method presented in this paper as a way of adapting TAS to the large-scale pretraining setting. Additionally, given the close relation between TAS and this work, it is quite plausible that the theory that inspired the TAS methodology may also help explain our empirical findings in the large-scale pretraining setting. Luckily, the TAS-B authors provide several nuggets of wisdom in their motivating remarks for us to draw from:

\begin{enumerate}
    \item They bring up the venerable ``cluster hypothesis'' \citep{clusterHypothesis,clusterRevisited}, which states that ``[i]t is intuitively plausible that the associations between documents convey information about the relevance of documents to requests''. In other words, in real-world datasets, items often naturally fall into groups of related query relevance.
    \item They motivate this concept of natural groups within training examples as the existence of \textit{topics}, and they argue that query-item pairs covering completely different topics will generally serve as uninformatively easy negative examples to one another during contrastive training.
\end{enumerate}

Given how neither of these arguments is tied to the small-batch, labeled-negative setting studied in the TAS-B paper \citep{tasb2021}, we happily draw from these in our synthesis, which we present below in \Cref{sec:synthesis}.

\subsection{Concept Two: The ANCE Perspective On Hard Negative Mining}\label{sec:hard_negative}

\begin{align}
    \mathcal{L}_\text{infoNCE} &= -\log \left( \frac{\exp(\text{similarity} (q, k_+))}{\sum_{i} \exp( \text{similarity}(q, k_i))} \right) \label{eq:infonce_base} \\
    \mathcal{L}_\text{infoNCE} &=  \log \left( \sum_{i} \exp( \text{similarity}(q, k_i) \right)  - \text{similarity} (q, k_+) \\
    \mathcal{L}_\text{infoNCE} &=  \operatorname{smooth max}_i  (\text{similarity}(q, k_i)) - \text{similarity} (q, k_+) \label{eq:infonce_rearranged} \\
    \mathcal{L}_\text{infoNCE} &\approx \operatorname{max}_i  (\text{similarity}(q, k_i)) - \text{similarity} (q, k_+) \label{eq:infonce_rearranged_approx}
\end{align}

Consider the ubiquitous InfoNCE contrastive loss function \citep{oord2019representationlearningcontrastivepredictive} given in \Cref{eq:infonce_base}. Rearranging the terms into \Cref{eq:infonce_rearranged,eq:infonce_rearranged_approx}, we see that this loss approximates discounting a query's maximum query-item similarity across all items by its similarity score associated with its positively labeled item. When the smooth maximum is tuned to be a close approximation to the maximum\footnote{This is often the case in practice, e.g. when we scale similarity scores by a high scaling constant (sometimes expressed alternatively as dividing them by a low temperature parameter).}, all items in each minibatch already scored as less relevant to the query than the positively-labeled item contribute close to nothing to the loss function and thus do close to nothing to train the model. Seeing this, we can intuitively grasp the motivation for the practice of hard negative mining: We must ensure each minibatch contains an ample number of \textit{informative} (i.e. difficult) negative examples, or else training progress will slow to a standstill.

Though several algorithms perform hard-negative mining, potentially the most relevant to this discussion is that of Approximate nearest neighbor Negative Contrastive Estimation (ANCE) \citep{xiong2020approximatenearestneighbornegative}, as the ANCE authors offer a motivation that is quite similar to the motivation given in the above paragraphs (though the ANCE authors' motivating disposition is more in-depth and mathematically rigorous). The method of ANCE is as follows: Embed the entire training dataset using a checkpoint of the model in training, construct an Approximate Nearest Neighbors (ANN) search index from these embeddings, and then construct new training minibatches by augmenting randomly sampled query-item pairs with the set of negative items that the ANN index retries as most relevant to the query (i.e. the hardest negatives).

The Arctic Embed report \citep{merrick2024arctic} confirmed that fine-tuning with hard negative examples mined by their embeddings was critical to reaching SOTA performance on the MTEB Retrieval benchmark \citep{muennighoff2023mtebmassivetextembedding}, though this work used only a simplified, non-iterative version of ANCE that leveraged a single fixed text embedding model checkpoint to construct training batches once. The Arctic Embed report also found evidence that imposing an upper threshold on negative similarity, i.e. mining ''hard but not-too-hard`` negatives, worked better than mining the hardest examples.

Taken together, the works of ANCE and Arctic Embed suggest that the ideal contrastive training minibatch should be carefully curated via ``mined'' examples rather than randomly sampled, avoiding both non-active negative examples and too-active negative examples and teaching the model something useful in all contrastive pairings.

\subsection{A Possible Synthesis: Topic-Aligned Embeddings And A Triangle Inequality Thought Experiment}\label{sec:synthesis}

\begin{figure}
	\centering
    \begin{subfigure}[t]{0.5\textwidth}
        \centering
        \includegraphics[width=0.6\linewidth,trim={8cm 8cm 0 0},clip]{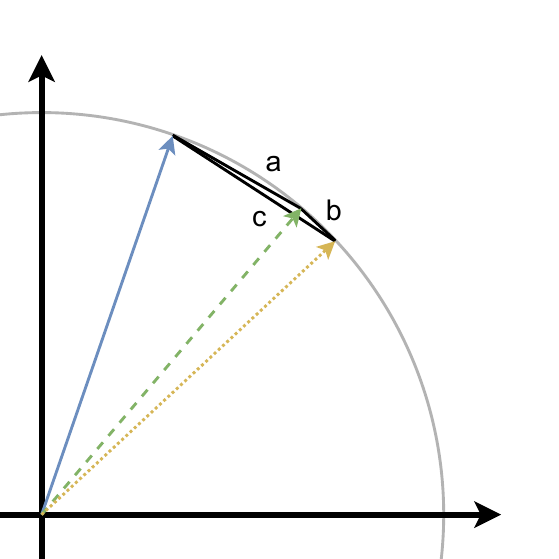}
        \caption{Guaranteed similarity.}
        \label{fig:triangle_close}
    \end{subfigure}%
    ~ 
    \begin{subfigure}[t]{0.5\textwidth}
        \centering
        \includegraphics[width=0.6\linewidth,trim={8cm 8cm 0 0},clip]{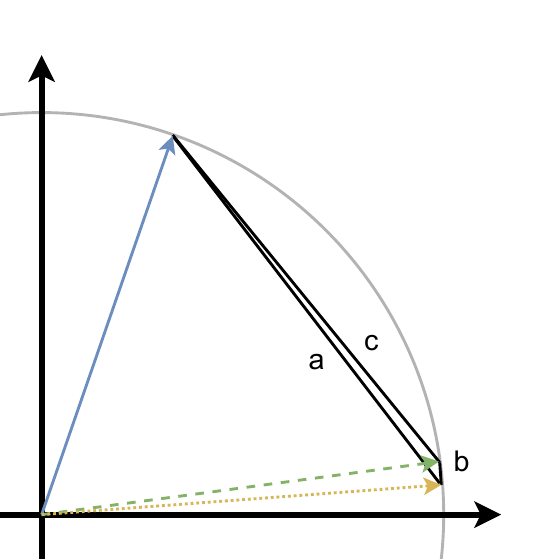}
        \caption{Guaranteed dissimilarity.}
        \label{fig:triangle_far}
    \end{subfigure}
	\caption{The triangle inequality $|a - b| \leq c \leq a + b$ guarantees that for any pair of similar vectors, a third vector that is similar to one of them cannot be too dissimilar to the other, while a third vector that is dissimilar to one cannot be too similar to the other.}
\end{figure}

Having discussed the relevant concepts of TAS-B and ANCE, we now attempt a synthesis of the ideas above into an explanation of the efficacy of clustering on embeddings. Let us consider the case of a dataset containing items that fall under various themes, e.g. MSMARCO, which covers a broad array of topics including health, finance, and technology. Let us further consider training an embedding model on this dataset to produce unit-norm vectors for vector similarity search. Let us further, for a moment, imagine a stage of training in which the model's embeddings are (1) well-trained enough to score most randomly sampled negative items substantially lower than the labeled positive item for most queries, and (2) have become geometrically well aligned with many of the semantic topics present in the data so that same-topic items cluster tightly in embedding space.

Given assumption (1), the motivating principles of hard negative mining from \Cref{sec:hard_negative} suggest that continuing to randomly sample in-batch negatives is a path towards a learning plateau. Furthermore, the cluster hypothesis discussed in \Cref{sec:cluster_hypothesis} gives us an intuitive way of thinking about the inefficiency of random sampling -- most negatives will be drawn from different topics than the positive item, which intuitively is easier to trivially discriminate as being off-topic.

However, by assumption (2), we can go a step further and pose a geometric argument regarding the difficulty of in-topic vs. out-of-topic negatives. When vectors for same-topic items are embedded close to one another geometrically (e.g. the dashed green and dotted yellow vectors visualized in \Cref{fig:triangle_close}), and query vectors are also embedded similarly to their labeled items (e.g. the solid blue vector falling close to the dashed green vector in \Cref{fig:triangle_close}), the triangle inequality $|a - b| \leq c \leq a + b$ guarantees that other in-topic items will all have some bare minimum level of similarity to the query as a result of their similarity to one another, motivating their inclusion as in-batch negatives. The triangle inequality also guarantees that as long as the query vector falls far from at least one off-topic item, all off-topic items will express some bare minimum level of dissimilarity from the query likewise due to the geometric similarity of same-cluster embeddings, motivating their exclusion as in-batch negatives, as illustrated in \Cref{fig:triangle_far}.

\subsection{Reality Check: Does This Explanation's Assumptions Hold Up In Practice?}

Having considered the thought experiment above, let us ask ourselves how believable our two assumptions are in the real-world practice of large-scale contrastive pretraining. The efficacy of ANCE \citep{xiong2020approximatenearestneighbornegative} and other hard-negative mining works offer convincing support for assumption (1) to hold, and the plateauing improvement in NDCG@10 score for the run with un-stratified data in \Cref{fig:copied_figure} agrees with this theory as well. Similar empirical support for assumption (2) is found by examining real clustered-by-embedding samples in \Cref{sec:cluster_examples}, as we see that our intuitive notions about topics do indeed match up with geometrically-close clusters of embedding vectors (for example, we notice a ``quantitative question'' topic in query cluster 5 and ``dollar value price'' topic in passage cluster 3 in \Cref{sec:cluster_examples}).

However, despite the embedding-topic alignment seen in \Cref{sec:cluster_examples}, the intra-cluster average-case similarity scores given in \Cref{tab:clusters} (in particular those in \Cref{tab:clusters_doc}) suggest that the clusters experimentally studied in this work may not be tightly packed enough to make the triangle inequality a convincing bound on negative hardness for all in-cluster negatives. In other words, assumption (2) may hold to a degree, but not so strongly that the triangle inequality guarantees that all same-cluster in-batch negatives will be usefully hard for the model.

A more granular clustering approach, e.g. the 2,000 clusters of \citep{tasb2021} or the one-cluster-per-minibatch case studied in \citep{cho2024minibatch}, may make for substantially denser clusters, but this would not explain the empirical benefit observed even at the $k=10$ clusters studied in this work. Though this appears to be a disagreement between theory and evidence, however, the disagreement only invalidates the argument for \textit{all} in-cluster negatives to be hard negatives. Since effective training does not require \textit{every} in-batch negative to be active for \textit{every} query, it still appears plausible that the principles discussed in \Cref{sec:synthesis} apply to the experiment at hand.

\section{Limitations And Alternatives}

\subsection{A Need For Curriculum Learning}

We discussed that increasing negatives is a compelling feature for the phase of training in which the model has already learned to generally score ``easy'' negatives as being less relevant than labeled positives. However, this is not necessarily the case at the beginning of training. While it is unclear that using clustered data at the beginning of training would be \textit{less} useful than un-clustered data, neither our theoretical arguments nor the empirical evidence of \Cref{fig:copied_figure} (where source-stratification does not improve performance in the first few hundred steps) suggest any benefits from this approach until the model has reached a baseline level of training.

Additionally if at some point a model is capable of correctly ranking the labeled items for most queries above those of nearly every other item in the dataset, even if a cluster contains all the remaining hard negatives, randomly sampling a relatively small minibatch from a relatively large cluster still risks the hard negatives not ending up in the minibatch. Thus a late-stage curriculum adjustment may be necessary in some cases, e.g. via shrinking cluster size (possibly to sub-batch-sized clusters) or switching over to fine-tuning-style training with hard-negative mining once the model reaches that point.

\subsection{Why Not Just Hard Negative Mine For Pretraining?}\label{sec:why_not_hard_negative}
\Cref{sec:synthesis} is in some ways an argument that clustering can approximate hard negative mining, which raises the question ``why not directly hard negative mine during pretraining?'' The primary answer here is one of efficiency: If we want more than one hard negative per query, hard-negative mining will require more than one negative item per query in the minibatch. Thus to train on each query, we must pay not just twice the embedding cost (one positive and one negative item), but several times (one positive and many negatives) the embedding cost incurred by the standard large-scale pretraining recipe.

Though clustering avoids this cost, other possible mitigations may exist as well. For example, one could try including in-batch \textit{positive} queries during hard negative mining -- for every negative item mined, we would bring the corresponding positive query into the minibatch as well to return the query-to-item ratio to 1. However, just like clustering, such an alternative would not outright guarantee meaningfully hard negatives in the batch for every positive, so it is unclear whether the benefits of such an approach would exceed those of the clustering approach.

\section{Related Work}

We have already covered several related works above in some detail, including the Arctic Embed \citep{merrick2024arctic} and Nomic Embed \citep{nussbaum2024nomic} technical reports which examined the efficacy of the source stratification approach, the TAS-B \citep{tasb2021} method which applied a similar embed-and-cluster approach to constructing training examples, and the ANCE method \citep{xiong2020approximatenearestneighbornegative} which also uses a pretrained text embedding model to guide the construction of minibatches with challenging negative examples. 

In addition to these works which we already discussed at some length above, the recent work of \citet{cho2024minibatch} also addresses the challenge of constructing minibatches for large-scale contrastive training, though from a different perspective. \citet{cho2024minibatch} first prove that the only set of size-$B$ minibatches for which an optimized solution also optimizes the full-batch loss over all $N$ example pairs is the combinatorially exhaustive set of all ${N \choose B}$ batches. From this perspective, the authors then examine how one might do their best with a tractably small set of minibatches, ultimately attacking the simpler problem of partitioning $N$ data points into $N / B$ batches and arriving at a familiar-looking embed-and-cluster approach. However, the methodological similarity to this work is somewhat limited given the fact that the authors seek to directly create minibatches (i.e. their concept of ``cluster'' and ``minibatch'' are one and the same) instead of thinking in terms of ``topics'' or ``sub-sources'' as we have. Additionally, instead of directly performing $k$-means clustering on embeddings, \citet{cho2024minibatch} apply a spectral clustering algorithm that uses both query and item embeddings and is motivated by a mathematical argument for maximizing a lower bound on total loss rather than an intuitive argument regarding natural semantic clusters in the data.

\section{Paths For Future Work}

In this work, we developed a deeper mental model of why the source stratification technique makes sense and provided a TAS-like algorithm that extends the technique and drives improvement on a real large-batch contrastive pretraining experiment. However, the core idea behind this paper -- that large-scale contrastive pretraining workloads can do better through clever minibatch construction -- extends well beyond what we have explored in these pages. Thus in place of a typical conclusion section, we instead opt for a discussion of several directions that future work may go from here.

\subsection{Tiny Dense Clusters}

This paper's work was motivated by source stratification and thus inherited a mindset of sampling multiple minibatches from each cluster. However, at large enough pretraining batch size, entire topics may fit into as few as a single batch, reversing the cluster-batch size relationship. TAS-B, for instance, created much smaller clusters averaging 200 queries per cluster\footnote{TAS-B created 2,000 clusters from 400,000 training queries, as well as a batch size of 32.} \citep{tasb2021}. One could imagine arguments for exactly one cluster per batch (e.g. maximizing the number of useful negatives for each query), as well as for multiple clusters per batch (some off-topic negatives might be helpful, especially early in training). A comparison to the in-batch positive concept suggested in \Cref{sec:why_not_hard_negative} would be elucidating as well.

\subsection{Smarter Clustering}

In a \Cref{sec:methodology} footnote, we mentioned that leveraging query and item information together during the clustering step might yield better results than clustering on either element alone. The spectral clustering method of \citep{cho2024minibatch} leverages information from both fields and proves useful in this setting, though a comparison to other non-spectral clustering approaches (such as simply concatenating query and item embedding vectors) may be a fruitful line of inquiry as well.

\subsection{Data Efficiency And Filtering}

In natural datasets, it is quite plausible that many queries may have few to no strongly relevant negative examples and thus cannot teach the model as much as examples for which more informative contrasts exist. Additionally, it is quite normal to have semantically redundant negative examples as well which demonstrate only a single important contrast despite incurring the computational cost of multiple embeddings. We believe that it may be possible to implement some data filtering into the cluster-and-sample approach presented in this paper and that doing so might lead to substantial efficiency improvements in real-world datasets.

For example, we saw that several clusters in \Cref{tab:clusters} were smaller and lower in density than the others (e.g. query clusters 2 and 9). While retaining outlier training examples may still help us train the embedding function to associate the query and item pair with one another, it may be beneficial to at least group these outliers into separate batches to allow the negatives of other batches more informative. The clustering approach used in this paper may naturally filter some outliers to their clusters, but understanding the degree to which this happens, the degree to which this helps model quality, and the degree to which this can be explicitly extended represent another compelling line for future work.

\subsection{Clustering Beyond Text}

Though this paper focused its experimentation on text embedding model training for information retrieval, applying a similar embed-and-cluster approach to vision or multimodal settings presents a natural extension.

\subsection{An Evolving Curriculum}

Just as ANCE \citep{xiong2020approximatenearestneighbornegative} mandates re-embedding the dataset and adjusting the construction of minibatches periodically, an analogous re-embed, re-cluster and re-sample approach could be applied in the pretraining phase as well. Effectively re-clustering more than once per epoch without excessively over-sampling certain examples also represents an interesting practical problem to explore. Additionally, the ideal cluster size may not be constant over training (e.g. \Cref{fig:copied_figure} suggests that source stratification matters more as training progresses, so perhaps the granularity of clustering should progress throughout training).

\section{Acknowledgements}

We thank Daniel Campos for his careful review and feedback on multiple drafts of this work.

\bibliographystyle{unsrtnat}
\bibliography{references}

\appendix

\section{Training Parameters}\label{sec:training_parameters}

\begin{itemize}
    \item 3 epochs at batch size 4,096
    \item Learning rate linear warmup for 50 steps, then linear decay from 4e-4 to 4e-5
    \item AdamW optimizer using PyTorch defaults for all parameters besides learning rate
    \item Gradient clipping to 1.0
    \item InfoNCE loss with temperature 0.02
    \item Truncated maximum passage length to 256 and maximum query length to 32
\end{itemize}

\section{Appendix: Examples From Clusters}\label{sec:cluster_examples}

Another motivation for why the clustering approach makes for better in-batch negatives is to simply look at the data. Examining a handful of random members from each query or passage cluster, we see that indeed there seems to be a higher semantic relevance between other members of the same cluster than there is between members of one cluster and members of another cluster. It is fairly easy to think of queries for which hard negatives would generally all stem from within the same cluster as the positive text.

\subsection{Query Cluster 1}
\begin{itemize}
  \item when is an international calendar
  \item was Peace Corps an executive order
  \item why is private ownership an important source of economic prosperity?
  \item why is business research important
  \item what is the standard deduction for married filing jointly
  \item how long to keep fsa documents
  \item what is a comprehensive deductible amount?
  \item can you use your paypal money to buy things
  \item what is competency and clia competency assessment?
  \item difference between a broker dealer and an ria
\end{itemize}

\subsection{Query Cluster 2}
\begin{itemize}
  \item driving distance boca raton to atlanta
  \item does a modem provide wifi
  \item when did cotogna open?
  \item what is the value of vintage rogers gas stove
  \item what is the height of mt si in north bend?
  \item amex stolen card phone number
  \item temperature in sydney, aus for march
  \item average monthly temperatures st. augustine fl
  \item weather in clayton
  \item how large is russia in sq miles
\end{itemize}

\subsection{Query Cluster 3}
\begin{itemize}
  \item what is a ghost account
  \item what is kitana
  \item what is a slug
  \item numbers in words form
  \item what is magix for
  \item what is filtration
  \item what is chai what species
  \item what does rfd mean in an
  \item what is marginal cost
  \item who is jeremy london
\end{itemize}

\subsection{Query Cluster 4}
\begin{itemize}
  \item can you make your legs longer naturally
  \item how does ph levels affect the color of rose petals?
  \item what chemical is found in apple seeds
  \item how long after power washing can i stain
  \item how long to cook a boiled egg
  \item what foods have algin
  \item which part of goat meat is best
  \item does thyme tea give insomnia
  \item how to bar b que beef ribs
  \item what to clean laminate floors with
\end{itemize}

\subsection{Query Cluster 5}
\begin{itemize}
  \item price of lic
  \item how much is a new alternator cost
  \item cost crusher tree removal
  \item average costs of bridge
  \item gpa requirements for baylor
  \item iceland average income
  \item home construction cost per square foot
  \item what should sugar cost
  \item cost of hemerroid
  \item cost of attending university of alabama
\end{itemize}

\subsection{Query Cluster 6}
\begin{itemize}
  \item what is lactobacilli
  \item what is pregnenolone steal
  \item do allergies cause bleeding
  \item what is msm vitamin supplement
  \item what type of medicine is cardiovascular
  \item can mri detect scar tissue
  \item which body part has primary responsibility for eliminating alcohol from the body
  \item what does the optic disk cause
  \item what are the characteristics of borderline personality disorder
  \item is codeine a vasodilator
\end{itemize}

\subsection{Query Cluster 7}
\begin{itemize}
  \item ministry define
  \item what is meaning of the surname dickinson
  \item what is meant by the term internal environment
  \item what is definition of knots
  \item what does the word jess mean
  \item what does the term ectopic mean
  \item what are felicitations?
  \item battery equalize charge definition
  \item earthworm crop definition
  \item impulsive definition
\end{itemize}

\subsection{Query Cluster 8}
\begin{itemize}
  \item where is cleveland
  \item who wrote when it comes to you, by john anderson
  \item who was melanie martinez's coach on the voice
  \item us navy roger william brown
  \item which president was nicknamed slick willie
  \item who sang fool me again
  \item when was march of dimes founded
  \item who was ted bundy's father
  \item are daddy longlegs venomous
  \item is flash faster than superman
\end{itemize}

\subsection{Query Cluster 9}
\begin{itemize}
  \item where is chrysler bldg located
  \item median home price in urbana, illinois
  \item what is riverview county
  \item population growth california
  \item what's the property tax rate for alameda county
  \item which county is miranda, ca
  \item what county is mount rainier in
  \item population irving texas
  \item where is winston salem nc
  \item what county is hickory creek texas
\end{itemize}

\subsection{Query Cluster 10}
\begin{itemize}
  \item the main function of the circulatory system is to \_\_\_\_\_\_\_\_\_\_.
  \item what kind of mollusk is an octopus
  \item what are critical limits in fish processing
  \item how deep should drain pipe be
  \item what is the iron deficiency problem in trees that requires iron chelate?
  \item why is argon unreactive
  \item what is non native species
  \item why is the nucleus important in eukaryotic cells
  \item where does bacteria live
  \item what is insulation made from
\end{itemize}

\subsection{Passage Cluster 1}
\begin{itemize}
\item August 3-6, 2017. The 2017 Pro Football Hall of Fame Enshrinement Week Powered by Johnson Controls kicks off with the annual Hall of Fame Game (Cardinals vs. Cowboys) on Thursday, Aug. 3.
\item Her sultry, powerful voice, her incredible legs, her time-tested beauty and her unforgettable story all contribute to her legendary status. Tina Turner was born Anna Mae Bullock in Nutbush, in Haywood County, Tennessee, to Zelma Priscilla (Currie) and Floyd Richard Bullock. Her family were sharecroppers. Tina was raised in the segregated South.
\item Macy's Herald Square, originally known as the R. H. Macy and Company Store, is the flagship of Macy's department stores, located on Herald Square in Manhattan, New York City.
\item Buddy Fite was my guitar teacher and guitar hero during the late sixties. Buddy Fite is the greatest jazz guitarist in my opinion. His albums can still be fo... Buddy Fite was my guitar teacher and guitar hero during the late sixties. Buddy Fite is the greatest jazz guitarist in my opinion. His albums can still be found by googling or Ebay. A must for every Jazz guitar player. I will be putting up more of his tunes. This is a pure love of the man and his guitar playing.
\item From Wikipedia, the free encyclopedia. The Terracotta Army or the Terracotta Warriors and Horses is a collection of terracotta sculptures depicting the armies of Qin Shi Huang, the first Emperor of China.he figures include warriors, chariots and horses. Estimates from 2007 were that the three pits containing the Terracotta Army held more than 8,000 soldiers, 130 chariots with 520 horses and 150 cavalry horses, the majority of which remained buried in the pits nearby Qin Shi Huang's mausoleum.
\end{itemize}

\subsection{Passage Cluster 2}
\begin{itemize}
\item The intertidal zone (sometimes referred to as the littoral zone) is the area that is exposed to the air at low tide and underwater at high tide (the area between the low and high tide lines).This area can include many different types of habitats, including steep rocky cliffs, sandy beaches, or wetlands.he intertidal zone (sometimes referred to as the littoral zone) is the area that is exposed to the air at low tide and underwater at high tide (the area between the low and high tide lines).
\item As described in the chapter's Continuity and Change section, what method did Sigmund Freud use to encourage his patients to talk freely? Answer Selected Answer: Free association Correct Answer: Free association Question 6. This preview has intentionally blurred sections.
\item Here are some ways to classify coal (by grouping it with similar types of rocks): Because slag formed from lava-like melted rock, it's sort of like an igneous rock -- but because humans made the melt, it's not a true igneous rock. So, we made up a new classification for slag: We call it a pseudo-igneous rock (pronounced SUE-doe ig-NEE-us).
\item The upper limb or upper extremity is the region in an animal extending from the deltoid region to the hand, including the arm, axilla and shoulder.Contents.ost of the large number of muscles in the forearm are divided into the wrist, hand, and finger extensors on the dorsal side (back of hand) and the ditto flexors in the superficial layers on the ventral side (side of palm). These muscles are attached to either the lateral or medial epicondyle of the humerus.
\item Scientists now know that independent assortment of genes occurs  during meiosis in eukaryotes. Meiosis is a form of cell division  that lowers the number of chromosomes in a parent cell by half to  make four reproductive cells called gametes.
\end{itemize}

\subsection{Passage Cluster 3}
\begin{itemize}
\item The median price for a house in the core Orlando market, which includes mostly Orange and Seminole counties, was \$181,900 in May that was up 10 percent from a year earlier and 4 percent from a month earlier, according to a report released Monday by Orlando Regional Realtors Association.
\item Cost of wisdom teeth removal - Extraction. As of 2017, our cost range from \$200 to \$500 per tooth for surgical wisdom teeth removal. This includes the cost of local anesthesia and follow visits. On average, the patient can expect to spend about \$1,400 for four wisdom teeth removal. The cost depends on the nature of the surgical extraction.
\item 1 On average, peonies can cost anywhere from \$2 for a packet of seeds to as much as \$3 to \$6 or more for each stem. 2  A packet of 500 poppy flower seeds retails for \$4 to \$7. 3  A Japanese Tree Peony that is in full bloom can retail for \$27 to \$39 from local nurseries.4  Peonies are a common flower used in various bouquets. For future brides that want to use this flower in a bouquet, the average price can fall between \$2 and \$5 per stem. 2  For example, the site MyFlowerBuyer.com sells different types of peonies for anywhere from \$350 to \$425 as a wholesale price.
\item The average cost to install galvanized or aluminum gutters is approximately \$4 to \$9 per linear foot. There are also vinyl gutters which are much easier to install, and which run at roughly \$3 to \$5 per linear foot. Therefore, installing from 125 to 200 feet of gutters will cost \$1050-\$2400. These prices, however, tend to apply strictly to the DIY homeowners.
\end{itemize}

\subsection{Passage Cluster 4}
\begin{itemize}
\item Although the age of sexual consent in Japan is 13 years of age, prefecture law usually overrides federal law, raising the age up to 18. The legal age of consent for sex in Tokyo and Nagano is 13, not 18 like the rest of the country. Jersey. 16.
\item 1 Arrangements under Part X of the Bankruptcy Act avoiding bankruptcy. 2  A person who is insolvent may avoid bankruptcy by reaching an understanding with creditors for the satisfaction of their claims. 3  This can be done by way of a personal insolvency agreement under Part X of the Bankruptcy Act.eople who are declared bankrupt are also unable to apply for certain jobs especially when they are un-discharged bankrupts. It's very difficult, for example, to work as a company director and obtain a position in the financial services sector if you are declared bankrupt.
\item This year's deadline to file your personal income tax return is midnight on May 5  five days later than the usual Apr. 30 cutoff. You can thank that nasty Heartbleed computer bug for the extension, after it forced the Canada Revenue Agency to shut down its E-File system earlier in April.his year's deadline to file your personal income tax return is midnight on May 5  five days later than the usual Apr. 30 cutoff. You can thank that nasty Heartbleed computer bug for the extension, after it forced the Canada Revenue Agency to shut down its E-File system earlier in April.
\item Real-time processing can help banks deliver a blended multichannel experience. For example, consider a customer who has an opening balance of \$250. That customer today deposits \$750 through the ATM. In a near real-time environment, the customer must wait up to a day or two for the bank processes that deposit before the funds become available. In a real-time processing system, that \$750 deposit would be cleared promptly, allowing the customer to make a transaction using the banks debit card, whether online or through a smart phone.
\item The amount of income that is used to calculate an individual's or a company's income tax due. Taxable income is generally described as gross income or adjusted gross income minus any deductions, exemptions or other adjustments that are allowable in that tax year.
\end{itemize}

\subsection{Passage Cluster 5}
\begin{itemize}
\item 2. Beet juice. How it works: Beets are a good source of potassium -- and a good source of folate, both of which are important in regulating blood pressure. What's more, beets contain nitrate, which is converted into nitrites once ingested. Nitrites relax smooth muscle tissue and increase blood flow.
\item Coconut oil is one of those pantry items that can help a dog with bad breath. It doesnt just boost digestive, immune system, and metabolic functions  it also helps to combat canine bad breath. Put a lovin teaspoonful over your dogs food every single day, and youll soon sniff sweeter breath  plus dogs love the taste; for them, coconut oil is a sweet treat.
\item There are 40 calories in 1 10 Chiclets serving of Chewing Gum (Sugared). Calorie breakdown: 1\% fat, 99\% carbs, 0\% protein.
\item The refrigerator is a very good storage area for flour, but the use of a sealed container is even more important to prevent the flour from absorbing moisture as well as odors and flavors from other foods stored in the refrigerator. The freezer is usually the best location for long term storage.
\item The highest-protein fruit, guava packs more than 4 grams per cup, along with 9 grams of fiber and only 112 calories. With 600\% of your DV of Vitamin C per cup the equivalent of more than seven medium oranges! the tropical fruit should merengue its way into your shopping cart ASAP.
\end{itemize}

\subsection{Passage Cluster 6}
\begin{itemize}
\item 1 Place the meat in the oven and cook for 20 to 30 minutes at the high temperature. 2  Then lower the temperature to between 275°F and 325°F and roast until done (see doneness guidelines below). Don't cover the pan. 2  If you are using a meat thermometer (analog or digital), insert the probe into the center of the roast, being careful not to hit bone. 3  Place the meat in the oven and cook for 20 to 30 minutes at the high temperature.
\item Whats the Difference Between Paleo and Keto? In the end, the main difference between Paleo and keto is one of emphasis. Keto emphasizes being in the state of ketosis whereas Paleo emphasizes food quality. In practice, most folks on a Paleo diet eat a much higher amount of carbohydrates than those on a ketogenic diet.
\item To mount your SKS you will need the Kalinka Optics. Warehouse Universal AK/SKS/SVD Side Plate, Undrilled which can be found in the. Mounts \& Rings section at www.kalinkaoptics.com. Once you have attached the plate to. your SKS you can use any AK, SKS/SVD or Russian Small Arms mount including all of. our side mounts and the POSP/PSO series of scope. Mounting your SKS by drilling and. tapping a plate is the absolute only way to mount an SKS, using a receiver cover is.
\item 2. Under Computer name, domain, and workgroup settings click on Change Settings. Manage Settings. 3. Under the tab Computer Name find the Change  button and click it. Change Workgroup Name. 4.Under Member Of change the Workgroup name. Change Workgroup Name. 5. Then click on OK, then when prompted reboot your device.hanging the Workgroup in Windows 10. Follow the steps below to change the workgroup in Windows 10. 1. With the right mouse button click the Start icon and choose System. If you have a touch enabled device, click and hold the start button, then tap the System button. Right-click Start Â» System.
\item Lake Isabella, CA - Weather forecast from Theweather.com. Weather conditions with updates on temperature, humidity, wind speed, snow, pressure, etc. for Lake Isabella, California Today: Sunny intervals, with a maximum temperature of 52° and a minimum temperature of 36°. Moderate west wind with maximum gusts of 30 mph.
\end{itemize}

\subsection{Passage Cluster 7}
\begin{itemize}
\item Phone Number of Allergan Contact is  +1(800)-347-4500, +1(714)-246-4500 . Allergan is a pharmaceuticals company that was established in 1948. It is multi-specialty health care company that focuses on medical dermatology, urology, neuroscience, ophthalmic pharmaceuticals and eye care.
\item Ocean Springs High School. Ocean Springs High School is an IB-certified public high school in Ocean Springs, Mississippi, United States. The school serves students in grades 912 and is part of the Ocean Springs School District. Contents.
\item The total driving distance from FLL to MIA is 27 miles or 43 kilometers. Your trip begins at Fort Lauderdale-Hollywood International Airport in Fort Lauderdale, Florida. It ends at Miami International Airport in Miami, Florida. If you are planning a road trip, you might also want to calculate the total driving time from FLL to MIA so you can see when you'll arrive at your destination.
\item Kalispell, Montana. Kalispell is a city in, and the county seat of Flathead County, Montana, United States. The 2015 Census estimates put Kalispell's population at 22,052. The Kalispell Micropolitan Statistical Area has a population of 93,068 and it is the largest city and commercial center of northwest Montana. The name Kalispell is a Salish word meaning flat land above the lake.
\item TEXAS CITY, TX 77590. Amoco Federal Credit Union's routing number (the leftmost number on the bottom of a check) is 313189391. Sometimes, banks have multiple routing numbers for different branches or uses. Please make sure this is the correct routing number for your branch!
\end{itemize}

\subsection{Passage Cluster 8}
\begin{itemize}
\item Within eukaryotes, DNA replication is controlled within the context of the cell cycle. As the cell grows and divides, it progresses through stages in the cell cycle; DNA replication takes place during the S phase (synthesis phase).
\item Natural Habitat: The natural habitat of starfish spans right from the intertidal zone, i.e., the seashore which is exposed to the air during the low tide and goes underwater during the high tide, to the abyssal zone, which has a depth of roughly about 4000 - 6000 meters.
\item There is another defect in tires that can cause a vibration called loaded road force variance. The tires and pass a visual inspection, a radial and lateral runout test, be perfectly balanced, but still cause a vibration. This is due to the internal defects in the tire.
\item The lungs are a pair of spongy, air-filled organs located on either side of the chest (thorax). The trachea (windpipe) conducts inhaled air into the lungs through its tubular branches, called bronchi. The bronchi then divide into smaller and smaller branches (bronchioles), finally becoming microscopic.
\item Computed tomography (CT scan or CAT scan) is a noninvasive diagnostic imaging procedure that uses a combination of X-rays and computer technology to produce horizontal, or axial, images (often called slices) of the body.
\end{itemize}

\subsection{Passage Cluster 9}
\begin{itemize}
\item Aldosterone is a hormone released by the adrenal glands. It helps the body regulate blood pressure. Aldosterone increases the reabsorption of sodium and water and the release of potassium in the kidneys. This action raises blood pressure.
\item Stimulants are drugs that can increase alertness and awareness, usually for a short time only. Most stimulants have more side-effects than other drugs. Some are classified as illegal drugs, most can cause addiction.For this reason, most legal stimulants are only available on prescription.Stimulants act on the nerves: Stimulants cause more neuro transmitters to the synapse (this is the gap between different nerves).ome are classified as illegal drugs, most can cause addiction. For this reason, most legal stimulants are only available on prescription. Stimulants act on the nerves: Stimulants cause more neuro transmitters to the synapse (this is the gap between different nerves).
\item Superior vena cava syndrome (SVCS) occurs when a persons superior vena cava is partially blocked or compressed. The superior vena cava is a major vein in a persons body. It carries blood from the head, neck, upper chest, and arms to the heart. Cancer is usually the main cause of SVCS.
\item Diarrhoea is usually a symptom of gastroenteritis (a bowel infection), which can be caused by: 1  a virus  such as norovirus or rotavirus. 2  bacteria  such as campylobacter, Clostridium difficile (C. 3  parasites  such as the Giardia intestinalis parasite that causes giardiasis.
\item Regardless of the severity of the injury, follow these steps to immediately treat a burn: Flush the burned area with cool running water for several minutes. Call 911 for a severe burn (see below to learn if your burn is severe) Apply a burn ointment or spray for pain. Take ibuprofen or acetaminophen for pain relief if necessary.
\end{itemize}

\subsection{Passage Cluster 10}
\begin{itemize}
\item impending adjective [before noun].  used to refer to an event, usually something unpleasant or unwanted, that is going to happen soon: impending disaster/doom The player announced his impending retirement from international football.
\item Medical Definition of hepatobiliary. : of, relating to, situated in or near, produced in, or affecting the liver and bile, bile ducts, and gallbladder hepatobiliary disease the hepatobiliary system.
\item Unstintingly definition, to be frugal; get along on a scanty allowance: Don't stint on the food. They stinted for years in order to save money. See more.
\item Definition of hunter. 1  1a : a person who hunts gameb : a dog used or trained for huntingc : a horse used or adapted for use in hunting with hounds; especially : a fast strong horse trained for cross-country work and jumping. 2  2 : one that searches for something. 3  3 : a pocket watch with a hinged protective cover.
\item What is Baroque Music? What is baroque, and when was the Baroque period? Derived from the Portuguese barroco, or oddly shaped pearl, the term baroque has been widely used since the nineteenth century to describe the period in Western European art music from about 1600 to 1750.
\end{itemize}

\end{document}